\documentclass[10pt,twocolumn,letterpaper]{article}

\usepackage[pagenumbers]{iccv} % To force page numbers, e.g. for an arXiv version

\usepackage{amsmath}
\usepackage{amssymb}
\usepackage{multirow}
\usepackage{makecell}
\usepackage{amsfonts}
\usepackage{mathrsfs}  
\usepackage{booktabs}
\usepackage{graphics}
\usepackage{subcaption}
\usepackage{pifont}
\usepackage[T1]{fontenc}
\usepackage[accsupp]{axessibility}  % Improves PDF readability for those with disabilities.

\newcommand{\cmark}{\textcolor{black}{\ding{51}}} %
\newcommand{\xmark}{\textcolor{black}{\ding{55}}} %
\newcommand{\stdv}[1]{\scriptsize$\pm$#1}

\newcommand{\eo}{$\Delta_{EO}$}

\newcommand{\DFRVAL}{DFR$_\text{Tr}^\text{Val}$~}

\definecolor{iccvblue}{rgb}{0.21,0.49,0.74}
\usepackage[pagebackref,breaklinks,colorlinks,allcolors=iccvblue]{hyperref}

\def\paperID{8731} % *** Enter the Paper ID here
\def\confName{ICCV}
\def\confYear{2025}

\title{Controllable Feature Whitening for Hyperparameter-Free Bias Mitigation}

\author{
Yooshin Cho~~~~Hanbyel Cho~~~~Janghyeon Lee~~~~HyeongGwon Hong~~~~Jaesung Ahn~~~~Junmo Kim \vspace{2.5mm} \\ 
Korea Advanced Institute of Science and Technology (KAIST) \vspace{0.5mm} \\   
{\tt\small \{choys95, tlrl4658, wkdgus9305, honggudrnjs, jaesung2, junmo.kim\}@kaist.ac.kr}
}

\begin{document}
\maketitle
\begin{abstract}
 As the use of artificial intelligence rapidly increases, the development of trustworthy artificial intelligence has become important. However, recent studies have shown that deep neural networks are susceptible to learn spurious correlations present in datasets. To improve the reliability, we propose a simple yet effective framework called controllable feature whitening. We quantify the linear correlation between the target and bias features by the covariance matrix, and eliminate it through the whitening module. Our results systemically demonstrate that removing the linear correlations between features fed into the last linear classifier significantly mitigates the bias, while avoiding the need to model intractable higher-order dependencies. A particular advantage of the proposed method is that it does not require regularization terms or adversarial learning, which often leads to unstable optimization in practice. Furthermore, we show that two fairness criteria, demographic parity and equalized odds, can be effectively handled by whitening with the re-weighted covariance matrix. Consequently, our method controls the trade-off between the utility and fairness of algorithms by adjusting the weighting coefficient. Finally, we validate that our method outperforms existing approaches on four benchmark datasets: Corrupted CIFAR-10, Biased FFHQ, WaterBirds, and Celeb-A. 
\end{abstract}    
\section{Introduction}
\label{sec:intro}

Deep neural networks have shown impressive performance by capturing task-relevant statistical cues from well-curated training datasets~\cite{he2016deep,zagoruyko2016wide}. However, if training datasets are poorly curated, neural networks often rely on spurious cues that do not generalize well beyond the training distribution. Nevertheless, it is challenging to determine which statistical cues are beneficial for task performance; thus neural networks often fail when train dataset is highly biased (i.e., datasets in which a target attribute has a strong spurious correlation with a particular bias attribute). For example, if neural networks rely on spurious correlations to predict the target attribute (e.g., recognizing objects by relying on backgrounds or textures), the generalization capability of the neural networks is severely reduced~\cite{zhu2016object, ribeiro2016should}. Furthermore, previous studies have empirically demonstrated that neural networks tend to focus on \textit{easier concepts}~\cite{arpit2017closer,kim2019nlnl}, and over rely on such spurious correlations~\cite{nam2020learning, lee2021learning}.

To address the issues, several studies have been proposed~\cite{lee2021learning, nam2020learning, bahng2020learning,creager2019flexibly}. A common strategy is to enforce networks to learn representations that are independent to the specified bias attributes by incorporating fairness criteria (e.g., \textit{demographic parity}, \textit{equalized odds}, and \textit{equal opportunity})~\cite{kim2019learning, zhu2021learning, ragonesi2021learning, bahng2020learning}. Previous works have quantified the fairness criteria using statistical measures (e.g., mutual information, Hilbert Schmidt Independence Criterion, and Hirschfeld-Gebelein-R\'enyi coefficient) that represent the dependency between model predictions (or representations) and bias attributes. However, these measures are often analytically intractable or computationally expensive to estimate directly. Therefore, they employed neural networks to estimate the measures, and achieved fairness by adopting adversarial learning or regularization terms. However, it should be noted that adversarial learning can be easily unstable, and careful tuning of hyperparameters is required for regularization terms. Furthermore, it is difficult to evaluate whether the neural estimator precisely estimates the dependency during the min-max game. 

\begin{figure*}[!t]
    \centering    \includegraphics[width=0.99\linewidth]{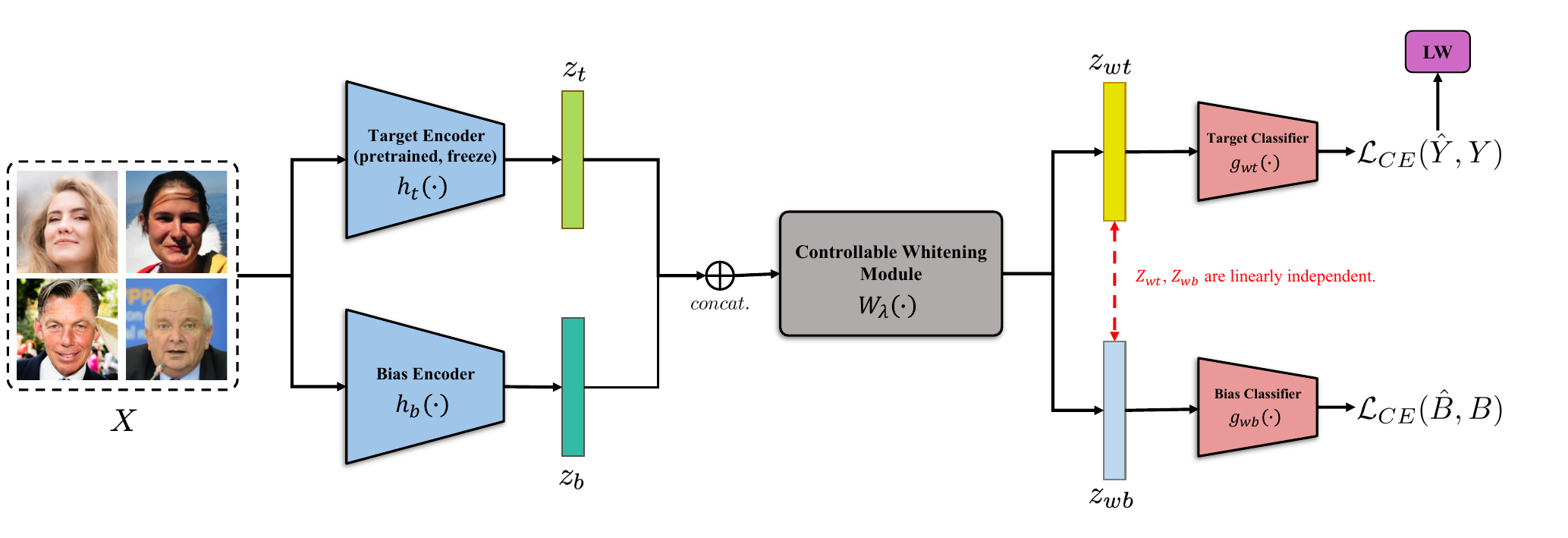}
    \vspace{-0mm}
    \caption{\textbf{Overview of proposed method.} $X$ is the mini-batch of the images that sampled from the biased training dataset. $h_t(\cdot)$ and $h_b(\cdot)$ are target and bias encoder, respectively. To reduce the dependency, we remove the linear correlation between the target feature $z_t$ and bias feature $z_b$ using the controllable whitening module $W_{\lambda}(\cdot)$, that can handle \textit{demographic parity} and \textit{equalized odds} by controlling the coefficient $\lambda$. Subsequently, the whitened target feature $z_{wt}$ and whitened bias feature $z_{wb}$ are linearly independent, while $z_{wt}$ is kept close to $z_t$ by the coupled Newton-Schultz iteration that utilizes the degree of freedom in the $\Sigma^{-1/2}$. Then, we train linear classifiers, $g_{wt}(\cdot)$ and $g_{wb}(\cdot)$, to predict the target attribute and bias attribute, respectively. $LW$ refers to the loss weighting. To make learning stable, we freeze $h_t(\cdot)$ which is pretrained on the same biased dataset. 
    }
    \label{fig:overall}
\end{figure*}

To overcome the limitations, we propose a simple yet effective feature whitening based approach that is robust to hyperparameter tunings and facilitates stable training. As noted in prior works, linear independence can be satisfied with the whitening transform that multiplies the inverse square root of the covariance matrix~\cite{huang2018decorrelated,huang2019iterative}. Although linear independence does not guarantee statistical independence, it ensures that one variable can not be estimated by a linear layer that takes other variables as inputs. This property allows our approach to function similarly to adversarial training, forcing the target representation to "forget" bias attributes in a linear regime. Furthermore, deep neural networks have empirically shown the capability to encode inputs into a representation that is linearly separable by a last linear classifier. Based on these insights, we estimate and eliminate the linear dependency between the target and bias features fed into the last linear classifiers, as illustrated in Figure~\ref{fig:overall}. Notably, we demonstrate that significant improvements in fairness can be achieved by removing only linear dependency, without involving analytically intractable higher-order dependencies, when the whitening transform is appropriately applied.

% Notably, despite the fact that we do not update the target encoder network, whitening the features successfully prevents the prediction of the target network from relying on the bias attributes without involving an unstable adversarial min-max game.

 In addition, we investigate the efficacy of the proposed method by evaluating \textit{demographic parity}, which is one of the most widely used fairness criteria. \textit{Demographic parity} requires the independence between the model predictions and the bias attribute. Experimental results confirm that our method effectively reduces \textit{demographic parity} as the bias feature is trained to be linearly classified according to the bias attribute. However, it is well known that enforcement of strict \textit{demographic parity} cripples the utility of the algorithm, particularly when the training dataset is highly biased, as it suppresses target-relevant information correlated with the bias attributes~\cite{hardt2016equality, seo2022information}. By contrast, \textit{equalized odds} means conditional independence between predictions and bias attributes given the target attributes; this does not conflict with learning target tasks regardless of the degree of the dataset bias~\cite{hardt2016equality}.

 To preserve task-relevant information while mitigating bias, we introduce a re-weighting strategy for covariance estimation. Specifically, we approximate a covariance matrix over the unbiased distribution by over-weighting the rare groups and under-weighting the predominant groups. In this paper, we denote the covariance matrix computed over the unbiased distribution (i.e., the target and bias attributes are independent) as the unbiased covariance matrix. Since \textit{equalized odds} and \textit{demographic parity} become equivalent in an unbiased distribution, whitening with the unbiased covariance matrix naturally promotes \textit{equalized odds}. Moreover, we empirically verify that whitening with the unbiased covariance matrix improves \textit{equalized odds}, and prevents the task-relevant information loss. However, we also confirm that whitening with the purely unbiased covariance can lead to over-fitting, due to the sample diversity imbalance between groups which can not be mitigated by re-weighting.

 To balance the trade-off between task-relevant information loss and over-fitting, we propose the Controllable Feature Whitening (CFW), which blends the unbiased and biased covariance matrices via a weighted arithmetic mean. By adjusting the weighting coefficient, we achieve a smoothly interpolated objective between \textit{demographic parity} and \textit{equalized odds}, enabling the model to mitigate inter-group performance disparities while simultaneously preserving the overall performance. Finally, we verify the efficacy of the proposed method by comparing the performance with existing methods on four benchmark datasets, Corrupted CIFAR-10~\cite{nam2020learning, ref_cifar}, Biased FFHQ~\cite{karras2019style, kim2021biaswap}, WaterBirds~\cite{sagawa2019distributionally}, and Celeb-A~\cite{liu2015deep}. Empirically, we observe that setting the weighting coefficient to 0.25 consistently yields strong performance across datasets, suggesting that our method can be considered hyperparameter-free in practice. Notably, we demonstrate that its effectiveness can be further enhanced when integrated with other existing methods which improve the quality of the representation.

\section{Related Work}
\label{sec:related}
\paragraph{Fairness with known biases.} Many approaches to improve fairness assume that bias attributes (e.g., gender, age, and race) are known during training. The most intuitive strategies are re-weighting and re-sampling~\cite{calders2009building, cadene2019rubi, roh2020fairbatch}, adjusting the training distribution to mitigate bias. However, re-weighting and re-sampling often lead networks to over-fitting due to the lack of the sample diversity of rare groups. To address the issues, data augmentation techniques such as BiasSwap~\cite{kim2021biaswap} and FlowAug~\cite{chiu2023better} generate counterfactual examples to enhance the sample diversity. However, data generation may adversely affect learning, depending on the quality of the generated data. Adversarial learning, another widely used approach, trains an auxiliary branch to predict bias attributes, forcing the target network to forget the bias information~\cite{kim2019learning,zhu2021learning}. Although adversarial learning based approaches have demonstrated strong performance, adversarial learning is inherently unstable and requires careful tuning of hyperparameters. In this paper, we apply a re-weighting strategy when estimating the unbiased covariance matrix, and demonstrate reduced over-fitting compared to standard loss re-weighting. Furthermore, we replace adversarial learning with feature whitening in a linear regime, and simplify the training by eliminating the need of additional hyperparameters.
\vspace{-3mm}
\paragraph{Fairness with unknown biases.} On the other hand, recent studies have focused on more challenging yet practical scenarios in which bias attributes are not provided during training. In such cases, models should infer bias attributes from datasets by leveraging properties of biased networks. Loss-based methods identify bias by detecting high-loss samples, assuming that they correspond to underrepresented groups~\cite{liu2021just, nam2020learning}. To amplify the loss difference between groups, the Generalized Cross Entropy (GCE) loss~\cite{zhang2018generalized} is widely utilized, and a committee of biased classifiers~\cite{kim2022learning} is introduced. However, these methods struggle to distinguish between truly biased and intrinsically difficult examples, making them sensitive to hyperparameter tuning and requiring a labeled validation set for effective calibration. Cluster-based approaches, such as Correct-n-Contrast~\cite{zhang2022correct}, seek to discover hidden bias attributes by grouping samples based on feature similarities, although such clustering may not always align with real-world biases. While fairness techniques that do not require explicit bias information have been increasingly studied and have shown even better performance than previous studies using the bias label, the absence of bias information can lead to limited generalization. Therefore, in this paper, we focus on the scenarios in which bias labels are provided, ensuring a more controlled approach to fairness.

\section{Preliminary}
\label{sec:preliminary}

\subsection{Whitening and Independence}
Whitening is a popular normalization technique which is widely adopted in various areas, including efficient optimization, domain adaptation, GAN, style transfer, and representation learning~\cite{cho2021improving, siarohin2018whitening, ermolov2021whitening, roy2019unsupervised}. It transforms the input features to have a zero mean and unit variance and removes the linear correlation between channels. It can be expressed by the following equation:
\begin{equation}
    \boldsymbol{\tilde{X}} = \boldsymbol{\Sigma^{-\frac{1}{2}}}\cdot (\boldsymbol{X} - \boldsymbol{\mu} \cdot \boldsymbol{1}^\top ), 
    \label{eq:whitening}
\end{equation}
where $\boldsymbol{X \in \mathbb{R}^{C \times N}}$ denotes the input features, $N$ denotes the number of inputs, and $C$ is the dimension size of inputs. $\boldsymbol{\Sigma} = \frac{1}{N} (\boldsymbol{X} - \boldsymbol{\mu} \cdot \boldsymbol{1}^\top ) \cdot (\boldsymbol{X} - \boldsymbol{\mu} \cdot \boldsymbol{1}^\top )^\top$ and $\boldsymbol{\mu} = \frac{1}{N} \boldsymbol{X} \cdot \boldsymbol{1} $ are the covariance matrix and mean vector of the input, respectively. Since the inverse square root of matrix is not unique (as multiplying any unitary matrix generates a valid inverse square root), many studies have been proposed such as ZCA-whitening~\cite{bell1997independent, kessy2018optimal}, Cholesky decomposition~\cite{dereniowski2003cholesky}, and Newton Schulz iterations~\cite{bini2005algorithms}. We employ the coupled Newton-Schultz iterations~\cite{higham2008functions, higham1986newton} which is known to be numerically stable and computationally efficient~\cite{ye2019network}.

Although zero covariance does not imply statistical independence between variables, it does ensure linear independence, which means that one variable can not be defined as a linear combination of the others (i.e., variables linearly unlearn each other). Moreover, if the variables follow a Gaussian distribution, a zero covariance implies statistical independence. Previous studies have analyzed that infinite-width neural network can be approximated as a Gaussian process by using the Central Limit Theorem~\cite{jacot2018neural, lee2017deep}. Although this does not rigorously match our work, we empirically demonstrate that whitening improves the fairness.

\subsection{Fairness Criterion}
\label{subsec:formulation}

\paragraph{\textbf{Problem Setup.}} Let $(X, Y, B) \in \mathcal{X} \times \mathcal{Y} \times \mathcal{B}$ denote the input data, target attribute, and bias attribute, respectively, sampled from the dataset $\mathcal{D}$. $\hat{Y} \in \mathcal{Y}$ denotes the algorithm prediction. We say that dataset $\mathcal{D}$ is biased towards $B$, when $Y$ and $B$ are not independent (i.e., $P(Y | \mathcal{D}) \cdot P(B | \mathcal{D}) \neq P(Y,B |\mathcal{D}) $). If a network is trained on a highly biased dataset, $B$ can be used as a shortcut to predict $Y$. Following~\cite{nam2020learning, lee2021learning}, we refer to data samples as \textbf{\textit{bias-aligned}} if $Y$ can be correctly predicted by relying on $B$, which are predominant in biased datasets. Conversely, we say data samples are\textbf{ \textit{bias-conflicting}}, if the $Y$ can not be predicted by relying on $B$, which are rare in biased datasets. If predictions, $\hat{Y}$, are highly biased toward $B$, the network poorly performs on \textit{bias-conflicting} samples. Therefore, the objective of this work is not only improving the overall performance, but also reducing the performance gap between the groups.
\vspace{-2mm}
\paragraph{\textbf{Fairness Criteria.}} Numerous fairness criteria have been proposed to measure the group fairness of algorithms, including \textit{demographic parity}~\cite{kilbertus2017avoiding, zhang2018mitigating}, \textit{predictive parity}~\cite{gajane2017formalizing}, \textit{equalized odds}, and \textit{equal opportunity}~\cite{hardt2016equality}. In the paper, we focus on two representative criteria: \textit{demographic parity} and \textit{equalized odds}, which are widely adopted to regularize training.

First, \textit{demographic parity} requires the difference of prediction probability between the bias groups to be zero. This can be expressed using the following equation:
\vspace{1mm}
\begin{align}
\label{eq:dp}
    P(\hat{Y}=y| B=b_1) = P(\hat{Y}=y | B=b_2) , 
\end{align}
\vspace{1mm}
for $\forall (y,b_1,b_2) \in \mathcal{Y} \times \mathcal{B}  \times \mathcal{B}$. This condition implies that the target prediction $\hat{Y}$ and bias $B$ are independent. Many successful approaches have been proposed to reduce the statistical dependency between $B$ and $\hat{Y}$ (or $B$ and the target features) to achieve fairness. However, because $B$ and $Y$ are highly correlated in a biased training dataset, removing the dependency between $B$ and $\hat{Y}$ over the biased dataset inevitably reduces the dependency between $Y$ and $\hat{Y}$~\cite{hardt2016equality} as well. In this paper, we quantify the degree of violation for \textit{demographic parity} as the following: 
\begin{align}
\label{eq:ddp}
    \Delta_{DP} =& \frac{1}{N_Y} \sum_{y \in \mathcal{Y}} \max_{b_1,b_2}\lvert P(\hat{Y}=y| B=b_1)\notag\\ &- P(\hat{Y}=y | B=b_2) \rvert,
\end{align} 
where $N_y$ is the number of the classes of $Y$.

On the other hand, \textit{equalized odds} requires the true positive ratio and false positive ratio over different bias groups to be the same, which can be expressed as the following equation:
\begin{align}
\label{eq:eo}
    P(\hat{Y}=y_1 | B=b_1,Y=y_2)\notag\\ = P(\hat{Y}=y_1 | B=b_2,Y=y_2),
\end{align}
for $ \forall (y_1,y_2,b_1,b_2) \in \mathcal{Y} \times \mathcal{Y} \times \mathcal{B}  \times \mathcal{B}$. \textit{Equalized odds} implies that $\hat{Y}$ and $B$ are conditionally independent given $Y$. Owing to the conditioning on $Y$, this criterion preserves the dependency between $Y$ and $\hat{Y}$, mitigating the risk of reducing predictive performance. In particular, \textit{demographic parity} and \textit{equalized odds} are equivalent if and only if $Y$ and $B$ are independent (i.e., dataset is unbiased). We quantify the degree of the violation for \textit{equalized odds} as the following: 
\vspace{1mm}
\begin{align}
\label{eq:deo}
    \Delta_{EO} =& \frac{1}{N_Y} \sum_{y \in \mathcal{Y}} \max_{b_1,b_2}\lvert P(\hat{Y}=y| B=b_1, Y=y) \notag\\ &- P(\hat{Y}=y | B=b_2, Y=y) \rvert.
\end{align}

\section{Methodology}
\label{sec:method}

\subsection{Training Debiased Classifier}
\label{subsec:debiasedcls}

In this section, we describe the proposed method in detail. The key component of our approach is the whitening module, which makes the features linearly independent between channels without requiring unstable adversarial learning or regularization terms. By leveraging this property, we just need to train the networks to predict the target and bias attributes using the different groups of the channels of the whitened features. As one group of whitened features is trained to linearly classify the bias attribute, the other group inherently becomes incapable of linearly encoding bias information. The overall framework of the proposed method is illustrated in Figure~\ref{fig:overall}.

Specifically, we train the network $f_t(\cdot): \mathcal{X} \rightarrow \mathcal{Y}$, which takes $X$ as the input and predicts the target attribute $Y$, over the biased dataset $\mathcal{D}_{b}$ using the standard cross-entropy loss. $f_{t} (\cdot)$ is composed of the encoder network $h_t (\cdot): \mathcal{X} \rightarrow \mathbb{R}^{M}$ and the linear classifier $g_t (\cdot): \mathbb{R}^{M} \rightarrow \mathcal{Y}$, where $M$ is the dimension size of the extracted target feature $z_t = h_t(X)$. Since, $Y$ and $B$ are highly correlated on $\mathcal{D}_{b}$, the network $f_t(\cdot)$ is likely to make predictions by relying on $B$ (i.e., $f_t(\cdot)$ shows great performance on the \textit{bias-aligned} samples, but poor performance on the \textit{bias-conflicting} samples). We refer this biased network $f_t(\cdot)$ as \textit{Vanilla} network. We bring the biased target encoder network $h_t(\cdot)$ from \textit{Vanilla} network and do not update, because satisfying fairness constraints, especially \textit{demographic parity}, is known to easily conflict with learning target tasks~\cite{hardt2016equality}. Moreover, previous works have observed that fine-tuning the last linear layer is sufficient to achieve fairness~\cite{tartaglione2021end, kirichenko2022last}. 

To mitigate the over-reliance of the pretrained \textit{Vanilla} network, we remove the bias information from the target feature $z_t$ using the feature whitening. We extract the bias feature $z_b=h_b(X)$ using another bias encoder network $h_b(\cdot) : \mathcal{X} \rightarrow \mathbb{R}^{M}$, which is trained to predict $B$. Then, the Controllable Feature Whitening (CFW) $W_\lambda(\cdot): \mathbb{R}^{2M} \rightarrow \mathbb{R}^{2M}$ takes the concatenated feature $z=[z_t;z_b]$ as the input, and performs whitening using the Eq~\ref{eq:whitening}. Detailed explanations of the whitening process will be provided in Section~\ref{subsec:covest}. Consequently, the whitened feature $z_w = W(z)$ satisfies the orthogonality between all channel pairs, and we split the $z_w$ into whitened target feature $z_{wt}$ and whitened bias feature $z_{wb}$. Then, the whitened target feature $z_{wt}$ and whitened bias feature $z_{wb}$ are linearly independent, while $z_{wt}$ is kept close to $z_t$ by the coupled Newton-Schultz iteration that utilizes the degree of freedom in the $\Sigma^{-1/2}$ in the whitening module. Owing to the linear independence, $z_{wt}$ and $z_{wb}$ can not be estimated by linear layers takes each other as the input.

Then, we train the linear target classifier $g_{wt} (\cdot): \mathbb{R}^{M} \rightarrow \mathcal{Y}$, which takes $z_{wt}$ as the input and predicts the $Y$. Similarly, we train the linear bias classifier $g_{wb} (\cdot): \mathbb{R}^{M} \rightarrow \mathcal{B}$, which takes $z_{wb}$ as the input and predicts $B$. For evaluation, we use $\hat{Y}=g_{wt}(z_{wt})$. The objective is composed of two terms: $\mathcal{L}_t(h_t, g_{wt}, X, Y) = \mathcal{L}_{CE}(g_{wt}(z_{wt}), Y)$ and $\mathcal{L}_b(h_b, g_{wb}, X, B) = \mathcal{L}_{CE}(g_{wb}(z_{wb}), B)$, where $[z_{wt};z_{wb}] = W([h_t(X);h_b(X)])$, and $\mathcal{L}_{CE}$ is the cross-entropy loss. Finally the objective function can be written as the follows:
\begin{align}
    \min_{g_{wt}}\mathcal{L}_t(h_t, g_{wt}, X, Y) + \min_{h_{b},g_{wb}} \mathcal{L}_b(h_b, g_{wb}, X, B).
\end{align}

\subsection{Covariance Estimation and Re-weighting}
\label{subsec:covest}

We improve fairness of the network by whitening with the biased covariance matrix which is estimated over the biased training dataset in Section~\ref{subsec:debiasedcls}. However, as we mentioned, removing the correlation between $\hat{Y}$ and $B$ can conflict with learning the target task when training dataset is highly biased. The problem is caused by the fact that $Y$ and $B$ are not independent in the training dataset. To avoid the problem, we propose the controllable covariance estimation using re-weighting. The re-weighting strategy is simple but effective method to mimic the statistics of unbiased dataset. We simply over-weight the rare groups (i.e., \textit{bias-conflicting} samples), and under-weight the common groups (i.e., \textit{bias-aligned} samples). The biased covariance matrix $\boldsymbol{\Sigma_{\text{b}}}$ and the unbiased covariance matrix $\boldsymbol{\Sigma_{\text{u}}}$ can be expressed as follows:
\begin{align}      
    \boldsymbol{\Sigma_{\text{b}}}=\sum_{y,b\in\mathcal{Y},\mathcal{B}}P(y,b|\mathcal{D}_\text{b}) \cdot \mathbb{E}_{\boldsymbol{X_c}\sim\mathcal{D}_\text{b}^{y,b}}[\boldsymbol{X_cX_c}^T],
\end{align}
\begin{align}
    \boldsymbol{\Sigma_{\text{u}}}&=\sum_{y,b\in\mathcal{Y},\mathcal{B}}P(y,b|\mathcal{D}_{\text{u}}) \cdot \mathbb{E}_{\boldsymbol{X_c} \sim \mathcal{D}_\text{b}^{y,b}}[\boldsymbol{X_cX_c}^T],
\end{align}
where $\mathcal{D}_{\text{b}}$ and $\mathcal{D}_{\text{u}}$ are the biased and unbiased distributions, respectively. $\mathcal{D}_\text{b}^{y,b}$ is the subset of $\mathcal{D}_\text{b}$ that contains samples with $Y=y$ and $B=b$. Thus, $P(y,b | \mathcal{D}_{\text{b}})$ can be obtained from the training dataset statistics. To ensure independence, we set $P(y,b | \mathcal{D}_{\text{u}}) = \frac{1}{N_Y \cdot N_B}$, where $N_Y$ and $N_B$ are the number of the classes of $Y$ and $B$, respectively. To obtain a more general expression of the covariance matrix for mixed distributions, we add a weight coefficient $\lambda \in [0,1]$ to compute the weighted arithmetic mean of the biased and unbiased covariance matrices as the following:
\begin{align}
    \boldsymbol{\Sigma_{\lambda}} &= \lambda \cdot \boldsymbol{\Sigma_{\text{u}}} + (1 - \lambda) \cdot \boldsymbol{\Sigma_{\text{b}}}.
\end{align}
Consequently, the proposed Controllable Feature Whitening (CFW) performs whitening with $\boldsymbol{\Sigma_\lambda}$ by following Eq~\ref{eq:whitening}. Setting $\lambda=0$, we can disable the re-weighting, and perform whitening with the biased covariance matrix, while increasing $\lambda$ gradually incorporates unbiased statistics.

\section{Experimental Results}
\label{sec:results}

To evaluate the efficacy of the proposed method, we compare the performance with existing methods on one constructed dataset (Corrupted CIFAR10~\cite{ref_cifar,nam2020learning}) and three real-world datasets (Biased FFHQ~\cite{kim2021biaswap}, WaterBirds~\cite{sagawa2019distributionally}, and Celeb-A~\cite{ref_celeba}). We conduct ablation studies to analyze 1) the contribution of each component in our method and 2) performance variations under different whitening transforms. Due to space constraints, 1) performance comparisons on the WaterBirds dataset, 2) implementation details, and 3) dataset descriptions are provided in Appendix. 

\vspace{-2mm}
\paragraph{Evaluation metrics.} Following previous studies, we report three types of accuracy: \textit{unbiased}, \textit{bias-conflicting}, and \textit{worst group}~\cite{tartaglione2021end,kim2022learning}. Detailed explanations and equations to compute the metrics are provided in Appendix. To sum up, we can evaluate the utility of algorithms with the \textit{unbiased test} accuracy, and the fairness of algorithms with the \textit{bias-conflicting} and \textit{worst-group} test accuracy. 

\begin{figure}[t]
    \centering
    \begin{subfigure}[b]{0.23\textwidth}
        \centering
        \includegraphics[width=\textwidth]{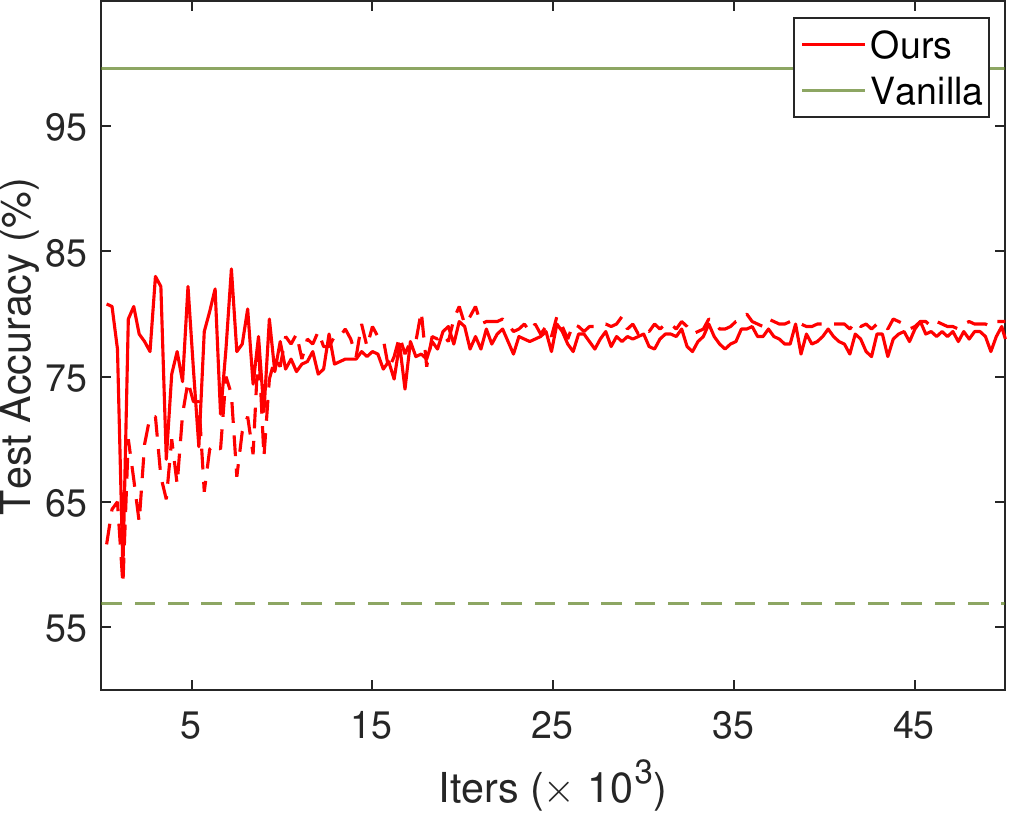}
         \subcaption{Test Accuracy}
         \label{subfig:test_acc_con,align}
    \end{subfigure} 
    \hfill
    \begin{subfigure}[b]{0.23\textwidth}
        \centering
        \includegraphics[width=\textwidth]{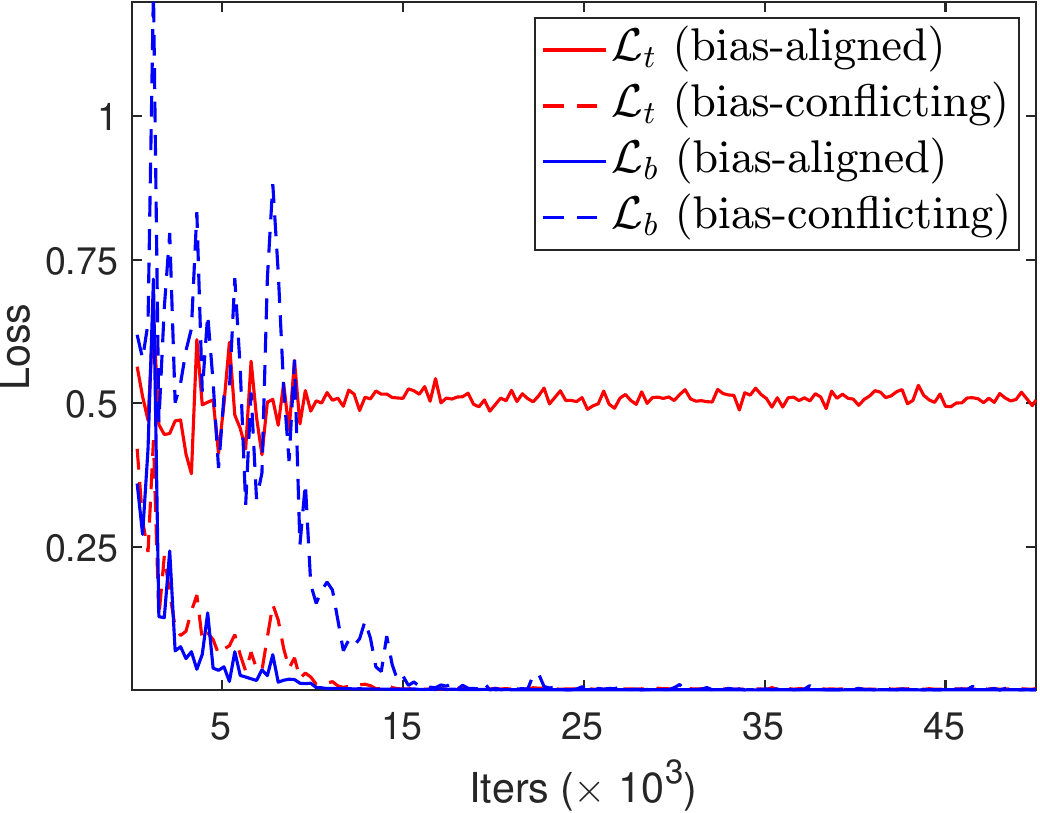}
        \subcaption{Training Loss}
        \label{subfig:train_loss_t,b}
    \end{subfigure}
    \vspace{-0mm}
    \caption{The solid lines and dashed lines in (a) are the test accuracy of the \textit{bias-aligned} and \textit{bias-conflicting} samples, respectively. For comparison, we also report the test accuracy of the \textit{Vanilla} network, which shares the target encoder $h_t$. (b) is the illustration of the training loss of $\mathcal{L}_t$ and $\mathcal{L}_b$ over the \textit{bias-aligned} and \textit{bias-conflicting} samples.}
    \label{fig:loss_tb,acc_con,align}
    \vspace{-2mm}
\end{figure}

\subsection{Controllable Feature Whitening}

To evaluate the efficacy of removing linear correlation between the target and bias features which are passed to the last linear layer, we conduct experiments on the bFFHQ dataset. In Figure~\ref{fig:loss_tb,acc_con,align}, we illustrate the test accuracy and the training loss of $\mathcal{L}_t$ and $\mathcal{L}_b$. As shown in Figure~\ref{subfig:test_acc_con,align}, the \textit{Vanilla} network, which is highly biased toward $B$, performs well only on the \textit{bias-aligned} samples (i.e., young women and old men). In contrast, the performance gap between two groups of our method is significantly reduced, and it indicates that the prediction of our method $\hat{Y}=g_{wt}(z_{wt})$ is not affected by the bias attribute. Despite the fact that we reuse the biased target encoder $h_t(\cdot)$ from the \textit{Vanilla} network without update, the proposed whitening module successfully removes the dependency between $z_{wt}$ and $B$ as $z_{wb}$ is trained to predict $B$. 

Furthermore, as shown in Figure~\ref{subfig:train_loss_t,b}, the training loss of $\mathcal{L}_t$ for the \textit{bias-aligned} samples does not converge, even though the \textit{bias-aligned} samples are majority in the dataset. By contrast, the training loss of $\mathcal{L}_t$ over the \textit{bias-conflicting} samples is well-converged, further indicating that the whitening module successfully removes the bias correlated information from $z_{wt}$ including the task-relevant information. To improve stability of training, we adopt the re-weighting strategy to $\mathcal{L}_t$ by under-weighting the loss of the \textit{bias-aligned} samples to mimic the loss of unbiased dataset. Empirically, it helps stabilizeining, and ablation studies will be provided in Section~\ref{subsec:abl}.

\begin{figure}[t]
    \centering
    \begin{subfigure}[b]{0.23\textwidth}
        \centering
        \includegraphics[width=\textwidth]{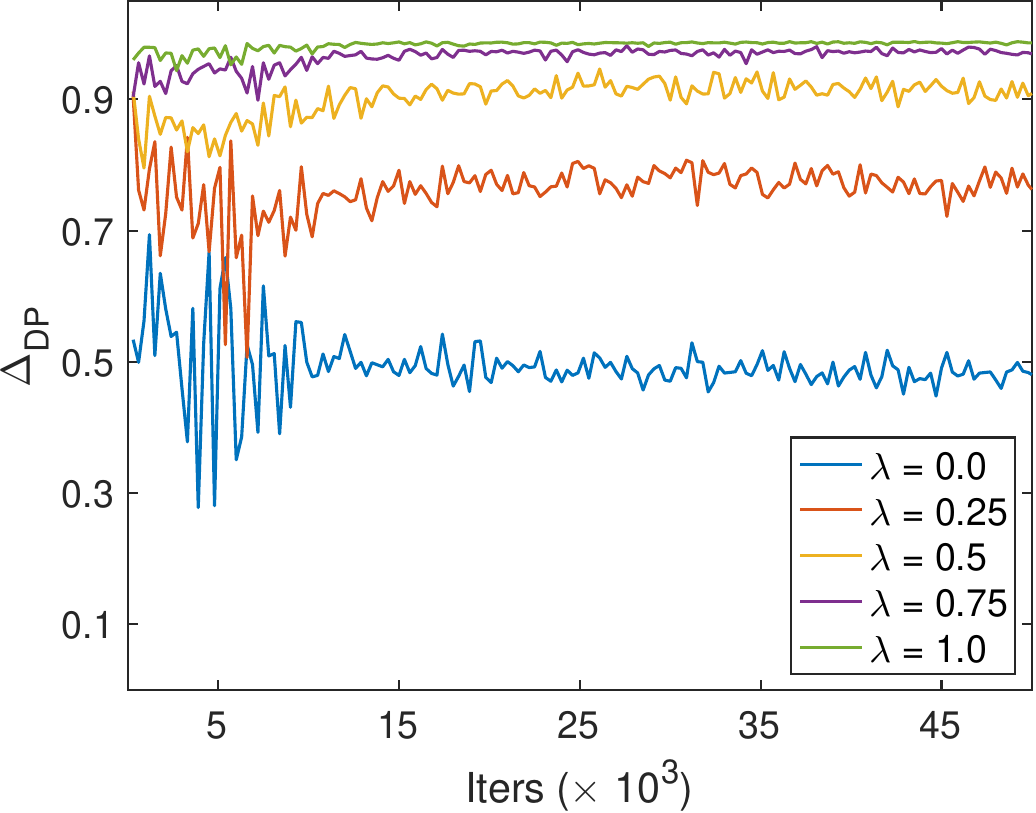}
        \subcaption{$\Delta_{DP}$}
    \end{subfigure}
    \hfill
    \begin{subfigure}[b]{0.23\textwidth}
        \centering
        \includegraphics[width=\textwidth]{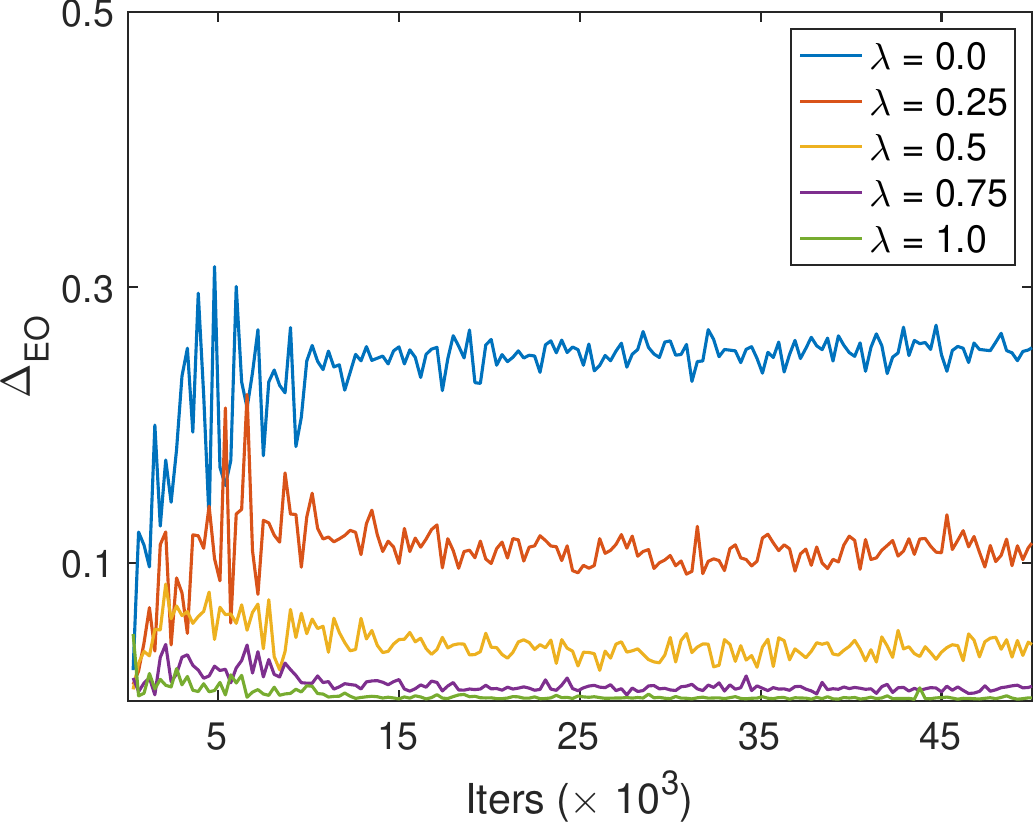}
         \subcaption{$\Delta_{EO}$}
    \end{subfigure} 
    \vspace{-1mm}
    \caption{Illustration of $\Delta_{DP}$ and $\Delta_{EO}$ with respect to training iterations. It shows that whitening with the unbiased covariance ($\lambda$=1) successfully regularizes $\Delta_{EO}$. By contrast, whitening with the biased covariance ($\lambda$=0) successfully regularizes the $\Delta_{DP}$.}
    \label{fig:fairness_measure}
\end{figure}

\begin{figure}[t]
    \centering
    \begin{subfigure}[b]{0.23\textwidth}
        \centering
        \includegraphics[width=\textwidth]{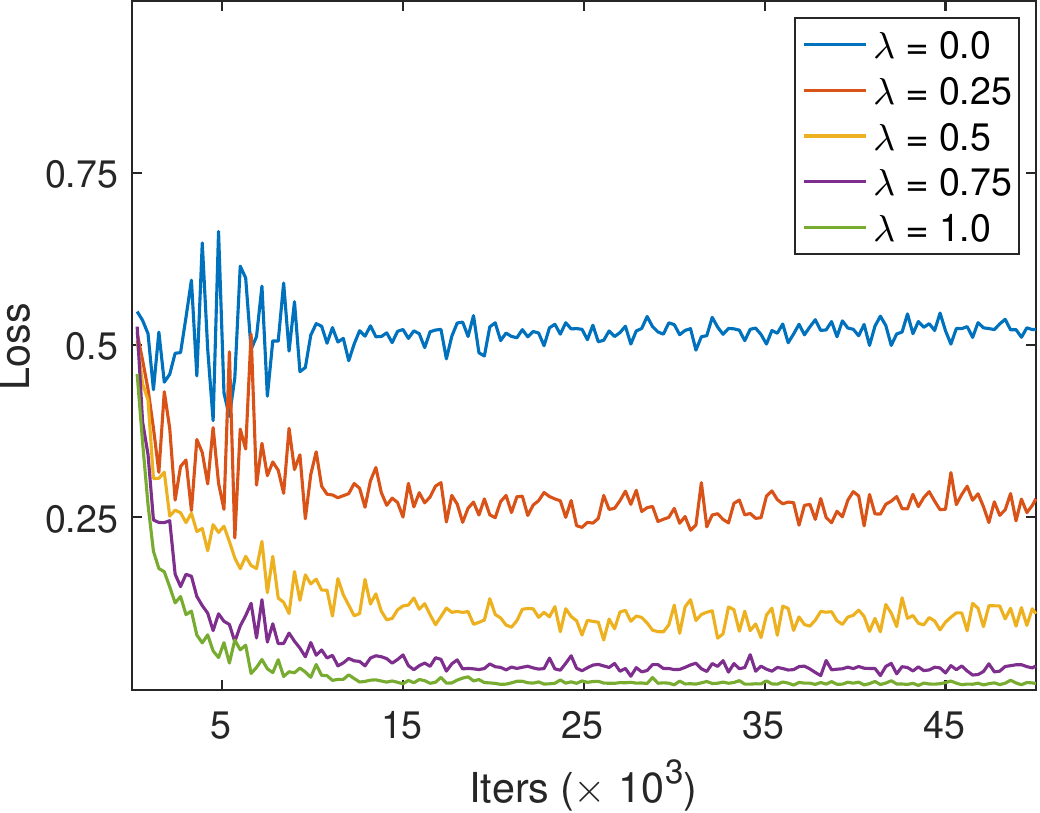}
        \subcaption{Training Loss (\textit{bias-aligned})}
        \label{subfig: train loss abl}
    \end{subfigure}
    \hfill
    \begin{subfigure}[b]{0.23\textwidth}
        \centering
        \includegraphics[width=\textwidth]{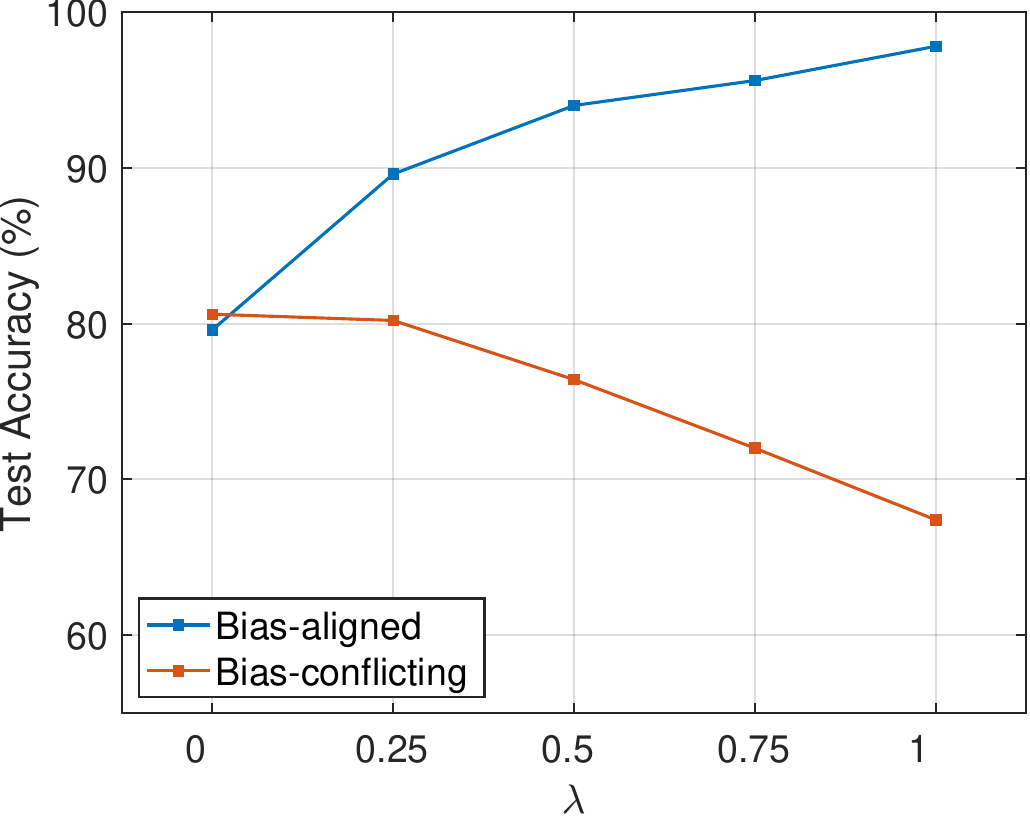}
         \subcaption{Test Accuracy}
         \label{subfig: test accuracy abl}
    \end{subfigure} 
    \caption{Illustration of training loss of \textit{bias-aligned} samples with respect to training iterations, and test accuracy with respect to $\lambda$. As $\lambda$ is getting larger, training loss of $\mathcal{L}_t$ on the \textit{bias-aligned} samples is converging better. However, the network tends to be over-fitting due to the sample diversity imbalance between two groups. 
    }
    \vspace{-2mm}
    \label{fig:loss_acc_lambda_abl}
\end{figure}

\begin{figure}[t]
    \centering
    \begin{subfigure}[b]{0.23\textwidth}
        \centering
        \includegraphics[width=\textwidth]{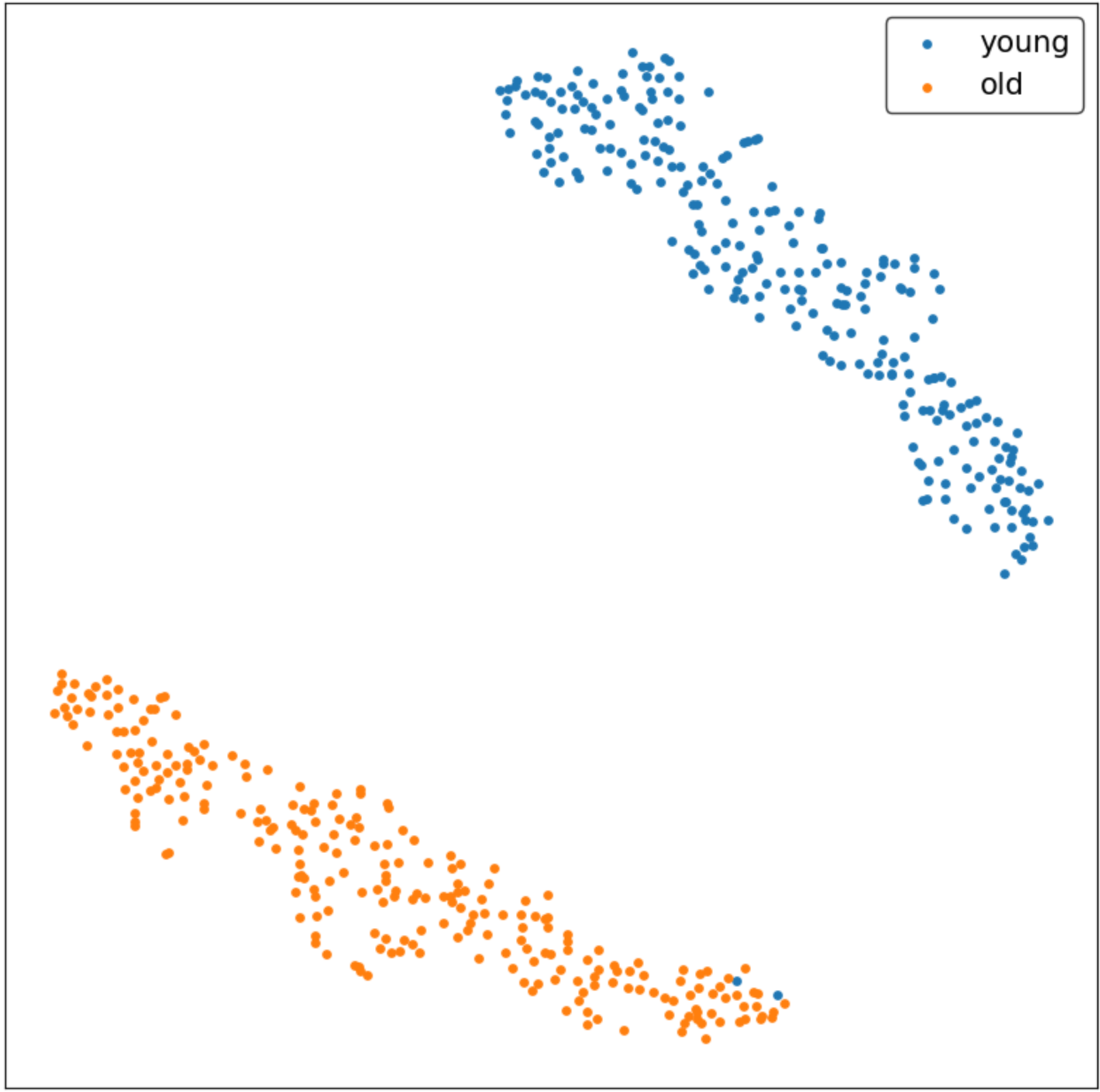}
        \subcaption{ \textit{bias-aligned} samples ($z_{t}$)}
        \label{subfig:t_align}
    \end{subfigure}
    \hfill
    \begin{subfigure}[b]{0.23\textwidth}
        \centering
        \includegraphics[width=\textwidth]{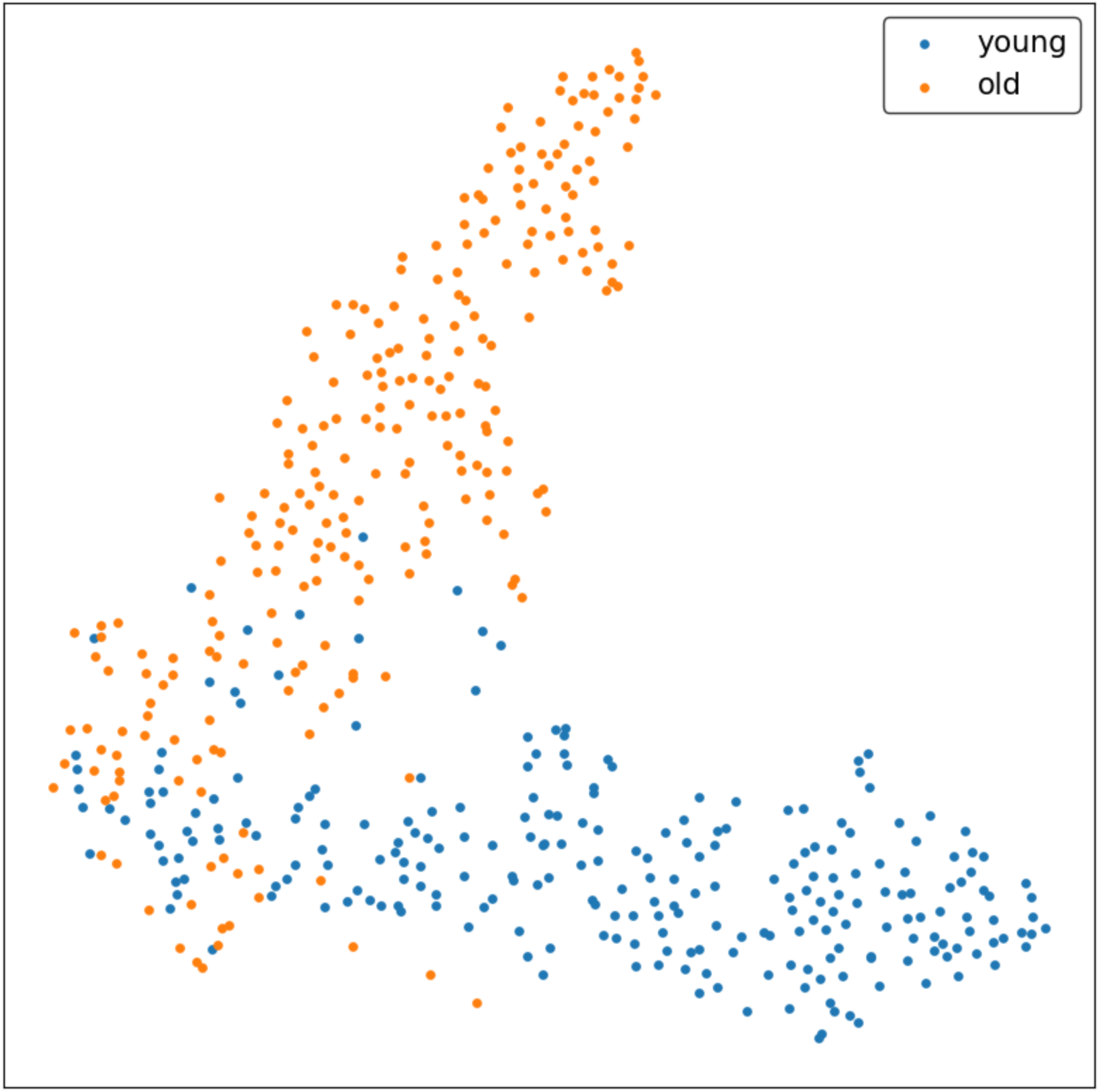}
         \subcaption{ \textit{bias-aligned} samples ($z_{wt}$)}
         \label{subfig:wt_align}
    \end{subfigure}
    \begin{subfigure}[b]{0.23\textwidth}
        \centering
        \includegraphics[width=\textwidth]{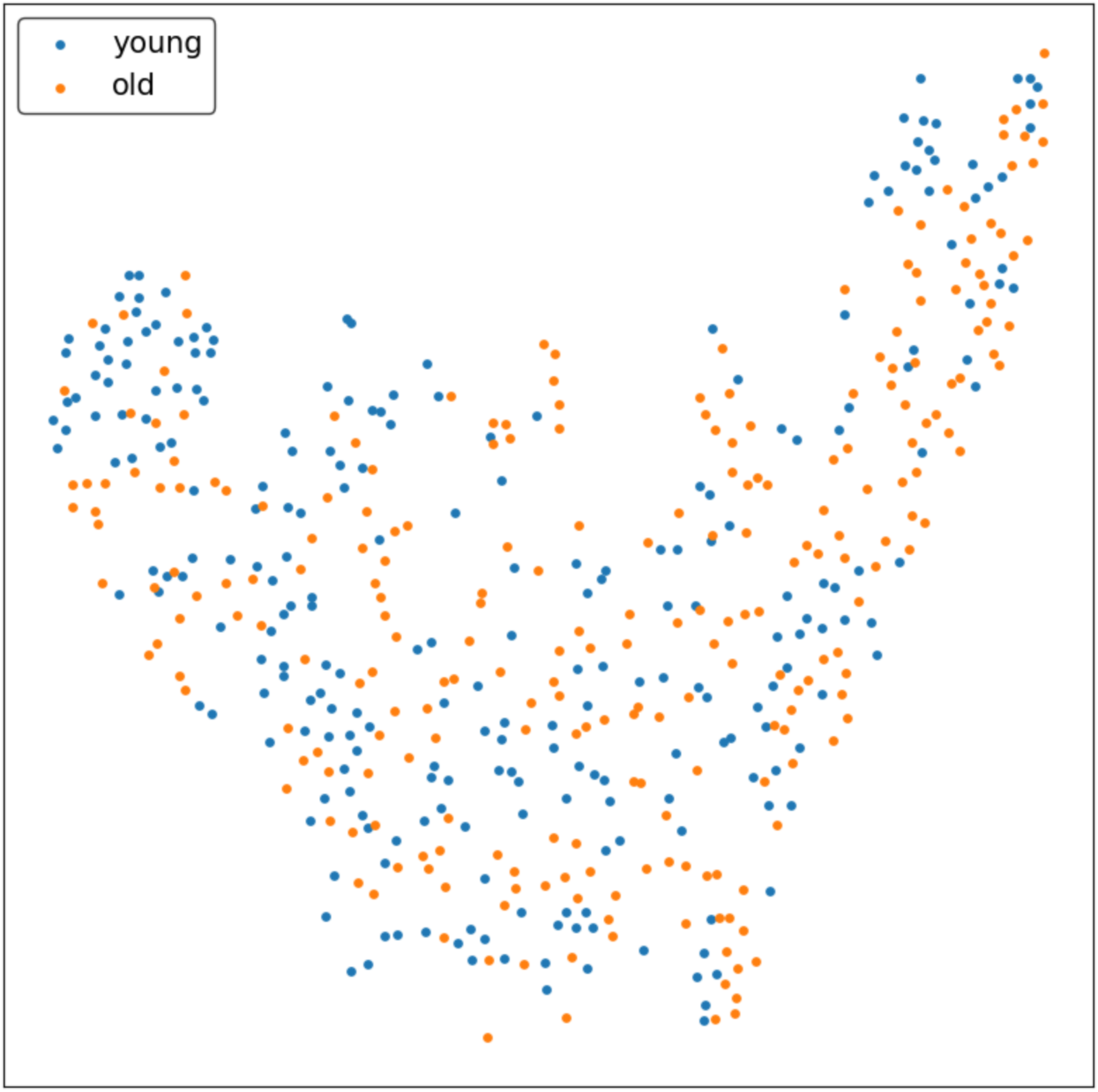}
        \subcaption{ \textit{bias-conflicting} samples ($z_{t}$)}
        \label{subfig:t_con}
    \end{subfigure}
    \hfill
    \begin{subfigure}[b]{0.23\textwidth}
        \centering
        \includegraphics[width=\textwidth]{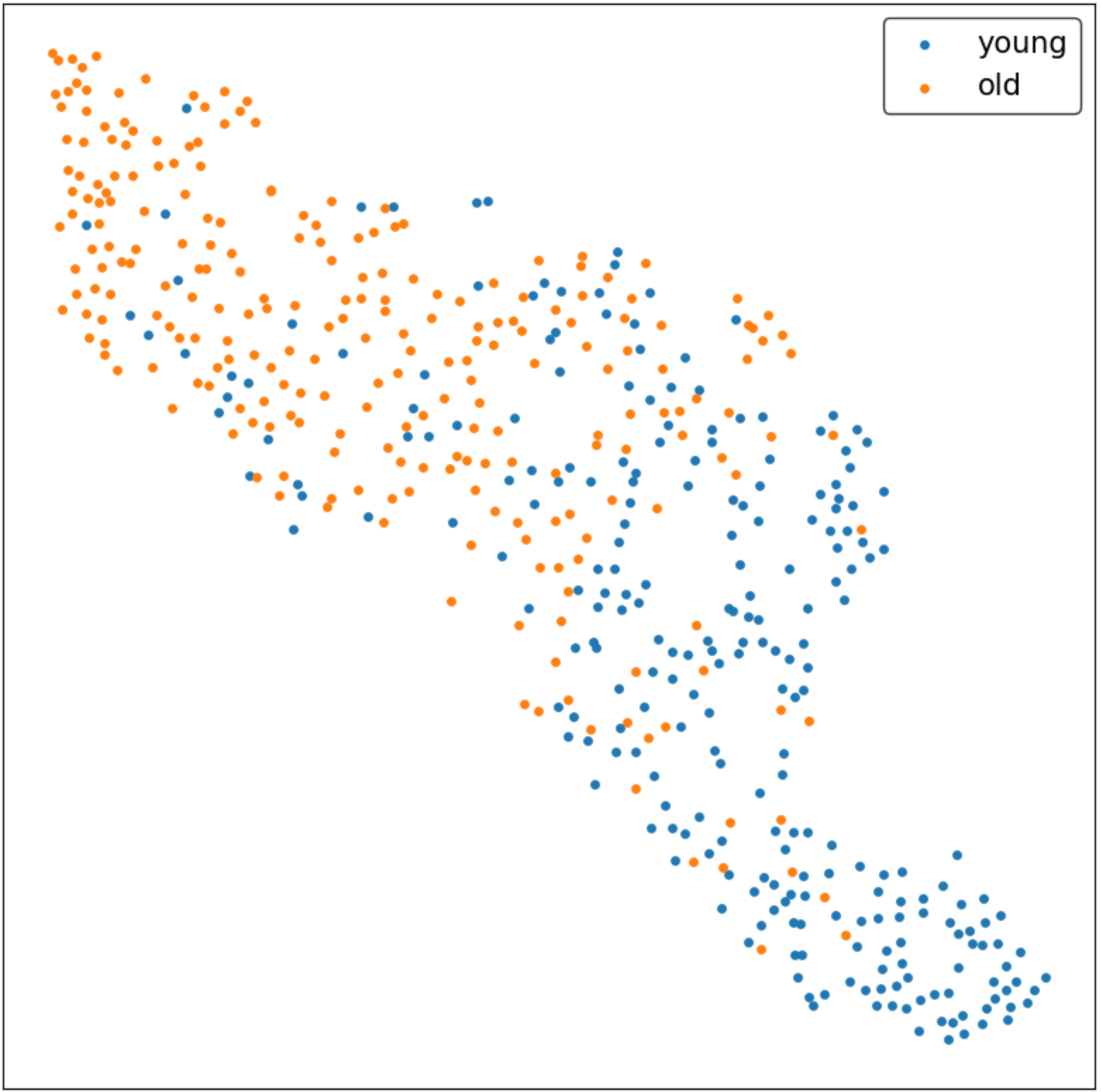}
         \subcaption{ \textit{bias-conflicting} samples ($z_{wt}$)}
         \label{subfig:wt_con}
    \end{subfigure}
    \vspace{-2mm}
    \caption{Illustration of the 2D projected target features $z_t$ and whitened target features $z_{wt}$ which are extracted with the biased FFHQ dataset. As we expected, $z_{wt}$ is consistently clustered according to the target attribute (old \& young) regardless of the groups, while $z_t$ is randomly mixed on \textit{bias-conflicting} samples.}
    \vspace{-2mm}
    \label{fig:tsne_bffhq}
\end{figure}

To preserve task-relevant information, we introduce the Controllable Feature Whitening (CFW), and demonstrate the efficacy by comparing the performance as varying the weight coefficient $\lambda$. In Figure~\ref{fig:fairness_measure}, we present $\Delta_{DP}$ and $\Delta_{EO}$ over the training dataset, computed by Eq~\ref{eq:ddp} and \ref{eq:deo}. The results allow us to empirically verify whether elimination of linear correlations through CFW can effectively regulate $\Delta_{DP}$ and $\Delta_{EO}$. As we expected, with $\lambda$ of $0$, we observe low $\Delta_{DP}$ and high $\Delta_{EO}$. It indicates that whitening with the biased covariance matrix focuses on removing dependency between $\hat{Y}$ and $B$ without conditioning on $Y$. By contrast, with $\lambda$ of $1$, we observe high $\Delta_{DP}$ and low $\Delta_{EO}$. It indicates that whitening with the unbiased covariance matrix removes the dependency between $\hat{Y}$ and $B$ conditioning on $Y$. These results confirm that removing linear correlations between target and bias features with CFW allows us to regulate smoothly interpolated objective between \textit{demographic parity} and \textit{equalized odds} by adjusting $\lambda$.

\begin{table*}[t]
\centering
\resizebox{0.75\textwidth}{!}{%
\begin{tabular}{lcccccc}
\toprule
\multirow{2}{*}{\makecell{Method}} &\multirow{2}{*}{\makecell{Bias Label}}   & \multicolumn{4}{c}{\makecell{Corrupted CIFAR-10}} & \multicolumn{1}{c}{\makecell{bFFHQ}} \\
\cmidrule(lr){3-6} \cmidrule(lr){7-7}
\multicolumn{1}{c}{\makecell{}} &
& 0.5\%
& 1.0\%
& 2.0\%
& 5.0\%
& 0.5\%
\\
\midrule
Vanilla &\xmark & {23.26}\stdv{0.29} & {26.10}\stdv{0.72} & {31.04}\stdv{0.44} & {41.98}\stdv{0.12} & {56.20}\stdv{0.35} \\ 
HEX~\cite{wang2019learning} &\xmark &  13.87\stdv{0.06} & 14.81\stdv{0.42} & 15.20\stdv{0.54} & 16.04\stdv{0.63} & 52.83\stdv{0.90} \\ 
Rebias~\cite{bahng2020learning} &\xmark &  22.27\stdv{0.41} & 25.72\stdv{0.20} & 31.66\stdv{0.43} & 43.43\stdv{0.41} & 59.46\stdv{0.64} \\ 

LfF~\cite{nam2020learning} &\xmark &  28.57\stdv{1.30} & 33.07\stdv{0.77} & 39.91\stdv{0.30} & 50.27\stdv{1.56} & 62.2\stdv{1.0} \\ 
DisEnt~\cite{lee2021learning} &\xmark &  29.95\stdv{0.71} & 36.49\stdv{1.79} & 41.78\stdv{2.29} & 51.13\stdv{1.28} & 63.87\stdv{0.31} \\ 
SelecMix+L (w/o GT)~\cite{hwang2022selecmix} &\xmark &  39.44\stdv{0.22} & 43.68\stdv{0.51} & 49.70\stdv{0.54} & 57.03\stdv{0.48} & 70.80\stdv{2.95} \\ \midrule
EnD~\cite{tartaglione2021end} &\cmark &  22.54\stdv{0.65} & 26.20\stdv{0.39} & 32.99\stdv{0.33} & 44.90\stdv{0.37} & 56.53\stdv{0.61} \\ 
LISA~\cite{yao2022improving} &\cmark & 32.71\stdv{1.09}&38.18\stdv{0.90}&44.15\stdv{0.39}&51.57\stdv{0.45}&64.20\stdv{0.53}  \\ 

SelecMix+L (w GT)~\cite{hwang2022selecmix} &\cmark & 37.02\stdv{1.05}&41.66\stdv{1.10}&48.35\stdv{0.99}&53.47\stdv{0.53}&75.00\stdv{0.53} \\ 
Ours+V &\cmark &  32.08\stdv{0.32} & 36.13\stdv{0.34} & 43.51\stdv{0.16} & 53.08\stdv{0.22} & 79.8\stdv{0.33} \\
Ours+S &\cmark &  \textbf{42.51\stdv{0.17}} & \textbf{46.87\stdv{0.12}} & \textbf{50.99\stdv{0.16}} & \textbf{59.05\stdv{0.23}} & \textbf{82.77\stdv{0.35}} \\ 
\bottomrule
\end{tabular}
}
\caption{Comparison of the unbiased test accuracy (\%) on Corrupted CIFAR10 and bias-conflicting test accuracy (\%) on bFFHQ. We compare the performance as varying the ratio of \textit{bias-conflicting} samples in the training dataset. We adopted target encoder networks pretrained with Vanilla and SelecMix for `Ours+V' and `Ours+S'. 
}
\vspace{-2mm}
\label{tab:c10,bffhq}
\end{table*}

\begin{table}[t]
\centering
\resizebox{0.99\columnwidth}{!}{
\begin{tabular}{lccccc}
\hline
Method   & Backbone & Bias Label     & \textit{Unbiased}    & \textit{Bias-conflicting} & \textit{Worst-G}   \\ \hline \hline
\multicolumn{6}{c}{Target attribute: \textit{BlondHair}}   \\
Vanilla  & Res18 (SL) & \xmark & 70.25\stdv{0.35}       & 52.52\stdv{0.19}   &     16.48    \\
LfF~\cite{nam2020learning} & Res18 (SL)  & \xmark & 85.43      & 83.40       & -     \\
LWBC~\cite{kim2022learning}   & Res18 (SL)  & \xmark& 85.1\stdv{0.6}      & 82.4\stdv{1.4}   &   76.6\stdv{4.6}      \\
Co-Ada~\cite{zhang2022contrastive} & Res18 (SL)  & \xmark & -    & -   &   78.37   \\
CM~\cite{basu2025mitigating} & Res18 (SL)  & \xmark & -    & -   &   81.61   \\
GroupDro~\cite{sagawa2019distributionally}   & Res18 (SL) & \cmark & 84.24      & 81.24   &    -      \\
CSAD~\cite{zhu2021learning} & Res18 (SL) & \cmark & 89.36     & 87.53       & -      \\
EnD~\cite{tartaglione2021end} & Res18 (SL)  & \cmark & \textbf{91.21}      & 87.45   &    -    \\
Ours    &  Res18 (SL) & \cmark & 88.40\stdv{0.40}       &  \textbf{88.07\stdv{0.16}}   & \textbf{84.03\stdv{0.19}} \\ \hline
Vanilla   & Res18 (SSL) & \xmark & 80.48\stdv{0.91}      & 66.79\stdv{2.20}    &    38.5\stdv{4.1}    \\
LWBC~\cite{kim2022learning}   & Res18 (SSL)  & \xmark& \textbf{88.90\stdv{1.55}}       & \textbf{87.22\stdv{1.14}}     &    \textbf{85.5\stdv{1.4}}   \\ \hline
Vanilla   & Res50 (SL) & \xmark&    -    & -   & 47.2             \\
JTT~\cite{liu2021just}   & Res50 (SL) & \xmark&    -     & - & 81.1            \\
CNC~\cite{zhang2022correct}   & Res50 (SL) & $\triangle$&    -  & -   &88.8\stdv{0.9}             \\
DFR~\cite{kirichenko2022last}  & Res50 (SL)  & $\triangle$&  91.3\stdv{0.3}     & -   &   88.3\stdv{1.1}      \\
FMD~\cite{chen2023fast}  & Res50 (SL)  & $\triangle$&  89.73     & -   &   87.15       \\
SSA~\cite{nam2022spread}   & Res50 (SL) & $\triangle$&    -    & -  &    89.8\stdv{1.28}      \\
GroupDro~\cite{sagawa2019distributionally}   & Res50 (SL) & \cmark&   -   & -  &    87.2     \\
LISA~\cite{yao2022improving}   & Res50 (SL) & \cmark&    -  & -  &89.30   \\
Ours    &  Res50 (SL) & \cmark & \textbf{92.33\stdv{0.13}}       &  \textbf{93.24\stdv{0.16}} & \textbf{91.02\stdv{0.17}} \\
\midrule \midrule
\multicolumn{6}{c}{Target attribute: \textit{HeavyMakeup}} \\ 
Vanilla   & Res18 (SL) & \xmark& 62.00\stdv{0.02}       & 33.75\stdv{0.28}  & -          \\
Vanilla   & Res18 (SSL)  & \xmark&66.30\stdv{1.15}      & 39.50\stdv{2.47}  & -        \\
LfF~\cite{nam2020learning} & Res18 (SL)  & \xmark& 66.20      & 45.48       & -     \\
LWBC~\cite{kim2022learning}   & Res18 (SSL)  & \xmark& 70.29\stdv{1.14}      & 51.28\stdv{5.74}  & -          \\
GroupDro~\cite{sagawa2019distributionally}   & Res18 (SL)  & \cmark& 64.88      & 50.24   &    -      \\
CSAD~\cite{zhu2021learning} & Res18 (SL)  & \cmark& 67.88     & 53.44      & -     \\
EnD~\cite{tartaglione2021end} & Res18 (SL)  & \cmark& \textbf{75.93}    & 53.70   &    -        \\
Ours    &  Res18 (SL)  & \cmark& 65.37\stdv{2.8}       &  \textbf{60.87\stdv{5.2}}   & \textbf{48.15 \stdv{6.42}} \\ \bottomrule 
\end{tabular}
}
\caption{Comparison of the \textit{unbiased}, \textit{bias-conflicting}, and \textit{worst-group} test accuracy (\%) on the Celeb-A. `SL' and `SSL' denote the adoptation of supervised and self-supervised models as the backbone network. 
\vspace{-3mm}
}
\label{tab:celeba}
\end{table}

To further analyze the effect of CFW, we visualize the training loss of $\mathcal{L}_t$ on the \textit{bias-aligned} samples and test accuracy evaluated on \textit{bias-aligned} samples and \textit{bias-conflicting} samples in Figure~\ref{fig:loss_acc_lambda_abl}. As shown in Figure~\ref{subfig: train loss abl}, the training loss of $\mathcal{L}_t$ in the \textit{bias-aligned} samples successfully converges as $\lambda$ increases. It indicates that the target information is well-preserved by re-weighting the covariance matrix. However, in Figure~\ref{subfig: test accuracy abl}, we empirically verify that the performance gap between the \textit{bias-conflicting} and \textit{bias-aligned} samples is growing with increasing $\lambda$. Although CFW demonstrates improved accuracy than the \textit{Vanilla} network regardless of $\lambda$, it seems that CFW with large $\lambda$ is prone to over-fitting due to the lack of the sample diversity of the \textit{bias-conflicting} samples. To avoid both target information loss and over-fitting, we empirically set $\lambda$ to 0.25 based on the results on the bFFHQ dataset. Notably, $\lambda$ of 0.25 is used in all experiments in Section~\ref{sec:results}, consistently yielding strong performance across datasets. This demonstrates that our method can be practically considered hyperparameter-free.

For qualitative evaluations, we visualize the 2D projections of the target features $z_t$ and whitened target features $z_{wt}$ using t-SNE in Figure~\ref{fig:tsne_bffhq}. These features are extracted from the bFFHQ dataset, and for better clarity, we separately visualize the projections for \textit{bias-aligned} and \textit{bias-conflicting} samples. A highly biased network only performs great on \textit{bias-aligned} samples, but a fair network consistently performs regardless of the groups. As shown in Figure~\ref{subfig:wt_align} and \ref{subfig:wt_con}, the whitened target features $z_{wt}$ are separable according to the target attribute (old \& young) regardless of the groups. On the other hand, as shown in Figure~\ref{subfig:t_align} and \ref{subfig:t_con}, the target features $z_{t}$ are only accurately separable on the \textit{bias-aligned} samples. It indicates that the proposed method successfully reduces the discrepancy between the groups, and prevents the network predictions from relying on the bias shortcuts.

\subsection{Classification Results}
\label{subsec:cls_results}

To compare the performance, we report the \textit{unbiased} and \textit{bias-conflicting} test accuracy on corrupted CIFAR-10 and bFFHQ, respectively, in Table~\ref{tab:c10,bffhq}. The proposed method demonstrates superior performance, and we note that its effectiveness further enhanced by adopting target encoder that pretrained with SelecMix. We denote our method employing target encoder networks pretrained with \textit{Vanilla} and SelecMix as `Ours+V' and `Ours+S', respectively. 

\begin{table*}[t]
\centering
\resizebox{0.8\textwidth}{!}{%
\begin{tabular}{cccccccccccccccccc}
\toprule
\multirow{2}{*}{Method } & \multicolumn{2}{c}{T=\textit{a} / S=\textit{m}} && \multicolumn{2}{c}{T=\textit{a} / S=\textit{y}} && \multicolumn{2}{c}{T=\textit{b} / S=\textit{m}} && \multicolumn{2}{c}{T=\textit{b} / S=\textit{y}}&& \multicolumn{2}{c}{T=\textit{e} / S=\textit{m}}&& \multicolumn{2}{c}{T=\textit{e} / S=\textit{y}}\\ \cmidrule[0.5pt]{2-3} \cmidrule[0.5pt]{5-6} \cmidrule[0.5pt]{8-9} \cmidrule[0.5pt]{11-12}  \cmidrule[0.5pt]{14-15} \cmidrule[0.5pt]{17-18}
& \eo  & Acc  && \eo   &  Acc   && \eo    &  Acc   &&  \eo   &  Acc  &&  \eo   &  Acc  &&  \eo   &  Acc \\ 
\cmidrule[0.5pt]{1-18} \morecmidrules\cmidrule[0.5pt]{1-18}
\textit{Vanilla} &  27.8 & 79.6 && 16.8 & 79.8 && 17.6 & \underline{84.0} && 14.7  & \underline{84.5} && 15.0 &83.9 && 12.7 & 83.8 \\ \cmidrule[0.5pt]{1-18} 
\textit{GRL}~\cite{raff2018gradient} & 24.9 & 77.2 && 14.7 &74.6 && 14.0 & 82.5 && 10.0 & 83.3 && 6.7 & 81.9 && 5.9 & 82.3  \\ 
\textit{LNL}~\cite{kim2019learning} & 21.8 & 79.9 && 13.7 & 74.3 && 10.7 & 82.3 && 6.8 & 82.3 && 5.0 & 81.6 && 3.3 & 80.3 \\ 
\textit{FD-VAE}~\cite{park2021learning} & 15.1 & 76.9 && 14.8 & 77.5 && 11.2 & 81.6 && 6.7 & 81.7 && 5.7 & 82.6 && 6.2 &84.0  \\ 
\textit{MFD}~\cite{jung2021fair} & 7.4 & 78.0 && 14.9 & \underline{80.0} && 7.3 & 78.0 && 5.4 & 78.0 && 8.7 & 79.0 && 5.2 & 78.0  \\ \cmidrule[0.5pt]{1-18}
\textit{SupCon}~\cite{khosla2020supervised} & 30.5 &\textbf{ 80.5 }&& 21.7 &\textbf{ 80.1} && 20.7 & \textbf{84.6} && 16.9 & 84.4 &&20.8 & 84.3  && 10.8 &84.0 \\ 
\textit{FSCL+}~\cite{park2022fair} & \underline{6.5} & 79.1 && \underline{12.4} & 79.1 && \textbf{4.7} & 82.9  && \underline{4.8} & 84.1  && \textbf{3.0 }& 83.4 && \textbf{1.6} & 83.5\\ \cmidrule[0.5pt]{1-18}
\textit{Ours} & \textbf{4.2} & \underline{80.3} && \textbf{10.4} & 79.4 &&  \underline{5.0} & 83.5  && \textbf{4.5} & \textbf{84.7} && \underline{3.4} &\textbf{ 85.3} && \underline{2.1} & \textbf{85.2}   \\\bottomrule
\end{tabular}%
}
\caption{Comparison of the top-1 unbiased test accuracy (\%) and equalized odds in various scenarios using Celeb-A. Here \textit{a}, \textit{b}, \textit{e}, \textit{m}, and \textit{y} respectively denote \textit{attractiveness}, \textit{bignose}, \textit{bag-under-eyes}, \textit{male}, and \textit{young}. On the other hand, T and S represent target and sensitive attributes, respectively.
}
\label{tab:celeba2}
\end{table*}

\begin{table}[ht]
\centering
\resizebox{0.95\columnwidth}{!}{%
\begin{tabular}{lccc}
\toprule
Method         & \textit{Unbiased} (\%)    & \textit{Bias-conflicting} (\%)  & diff. (\%)   \\ \midrule
Vanilla       & 77.96       & 56.87  & 21.09            \\
 + LW      & 78.5       & 56.2  & 22.03            \\
 + CFW, $\lambda=0.0$            & 80.0       &  \textbf{80.6}  & \textbf{-0.6}           \\ 
 + CFW, $\lambda=0.25$            & \underline{83.9}       &  78.1  & 5.8           \\ 
\midrule
 + CFW, $\lambda=0.25$ + LW    & \textbf{84.6}       & \underline{79.8}  & \underline{4.8}            \\ \bottomrule 
\end{tabular}%
}
\caption{Ablation study on each component of proposed method. We train and evaluate on the bFFHQ as varying the components. Best performing results are marked in bold, while second-best results are denoted with underlines.
}
\vspace{-2mm}
\label{tab:abl_method}
\end{table}
\begin{table}[ht]
\centering
\resizebox{0.95\columnwidth}{!}{%
\begin{tabular}{lccc}
\toprule
Method      & \textit{Unbiased} (\%)    & \textit{Bias-conflicting} (\%) & diff. (\%)  \\ \midrule
Vanilla       & 77.96       & 56.87  & 21.09            \\
ZCA       & 83.83       & 72.66  & 11.17            \\
CD      & 74.06       & 71.40  & \textbf{2.66}            \\
cNS, T=3    & \underline{84.50}     & \ 76.80  & 7.7           \\ 
cNS, T=7    & 83.80       & \underline{78.67} & 5.13            \\ 
\midrule
Ours: cNS, T=5    & \textbf{84.6}       & \textbf{79.8}  & \underline{4.8}            \\ \bottomrule 
\end{tabular}%
}
\caption{Ablation study on the method to solve the matrix inverse square root. We train and evaluate on the bFFHQ as varying the method. Best performing results are marked in bold, while the second-best results are denoted with underlines.
\vspace{-2mm}}
\label{tab:abl_wm}
\end{table}

 In Table~\ref{tab:celeba}, we compare the \textit{unbiased}, \textit{bias-conflicting}, and \textit{worst-group} accuracy of the proposed method with other existing methods on Celeb-A. For the \textit{bias-conflicting} and \textit{worst-group} accuracy, our method consistently outperforms other algorithms. For the \textit{unbiased} accuracy, our method shows comparable performance with the best result. In addition, the performance gap between the \textit{unbiased} and \textit{bias-conflicting} accuracy is significantly reduced with our method, indicating that our method improves the fairness without loss of the utility of algorithms. Additionally, in Table~\ref{tab:celeba2}, we compare the \textit{unbiased} test accuracy and $\Delta_{EO}$ of the proposed method with other baselines using bias label. The results confirm that our approach achieves both superior fairness and overall performance. Especially, we verify that our method shows significantly better performance than adversarial learning based methods (e.g., GRL and LNL).

\subsection{Ablation Study}
\label{subsec:abl}

To investigate the contribution of each component of our method, we conduct the ablation study by training ResNet-18 on the bFFHQ under different configurations. LW and CFW are abbreviation of loss weighting and Controllable Feature Whitening, respectively. As shown in Table~\ref{tab:abl_method}, we can verify that all components contribute to achieve the fairness while preserving the utility. Although, we verify that CFW without re-weighting demonstrates the smallest performance gap between two groups, we can optimize the \textit{unbiased} accuracy with the negligible \textit{bias-conflicting} accuracy drop by controlling the weight coefficient $\lambda$. Notably, LW does not improve the performance of the \textit{Vanilla} network, while it improves the performance of the proposed method. As we mentioned in Section~\ref{subsec:covest}, LW helps to stabilize the training by under-weighting the noisy gradients of the \textit{bias-aligned} samples. However, with \textit{Vanilla} network, training loss of both \textit{bias-aligned} and \textit{bias-conflicting} samples converge close to zero even without LW.

We further conduct the ablation study on the whitening module, training with ResNet-18 on the bFFHQ as varying the methods to solve the matrix inverse square root. We adopt three representative methods: ZCA-Whitening~\cite{bell1997independent, kessy2018optimal}, Cholesky Decomposition~\cite{dereniowski2003cholesky}, and coupled Newton-Schultz iterations~\cite{higham1986newton,higham2008functions}, which are denoted as ZCA, CD, and cNI, respectively. For cNI, there is the hyperparameter, iteration number T, which determines the number of iterations to approximate the inverse square root of the matrix. We use the cNI with T of 5, which is suggested by~\cite{huang2019iterative, ye2019network}. As shown in Table~\ref{tab:abl_wm}, the fairness and utility are generally improved regardless of which whitening modules are used. However, stochasticity of each modules are different. According to the previous works~\cite{huang2020investigation}, cNI demonstrates the most stable behavior, and it also outperforms other methods in our experiments. Moreover, we empirically verify that performance improvement is saturated with T of 5.

\section{Conclusion}
In this paper, we propose a novel framework, Controllable Feature Whitening (CFW), to mitigate over-reliance on spurious correlations by removing linear correlations between target features and bias features. Specifically, linear independence ensures that two features cannot be linearly predicted from each other. To enforce this, we whiten the features fed into the last linear classifier, effectively preventing model predictions from relying on bias attributes without requiring intractable modeling of higher-order dependencies. Additionally, we extend our method to achieve two fairness criteria, \textit{demographic parity} and \textit{equalized odds}, by re-weighting the covariance matrix. Although our approach assumes access to bias labels, it demonstrates consistently superior performance across datasets without requiring additional hyperparameter tuning. We validate its effectiveness by achieving state-of-the-art performance on four benchmark datasets: Corrupted CIFAR-10, biased FFHQ, WaterBirds, and Celeb-A. 

{
    \small
    \bibliographystyle{ieeenat_fullname}
    \bibliography{main}
}

\clearpage
\onecolumn

% \title{Supplementary Material \textit{for} \\ Controllable Feature Whitening for Hyperparameter-Free Bias Mitigation}
\begin{center}
    {\Large \bf Supplementary Material \textit{for}} \vspace{2mm} \\
    {\Large \bf Controllable Feature Whitening for Hyperparameter-Free Bias Mitigation}\\
\end{center}
\vspace{2em}
%%%%%%%%% AUTHORS - PLEASE UPDATE

\appendix

\section{Datasets}
\paragraph{Corrupted CIFAR-10} was proposed by~\cite{nam2020learning}, and constructed by corrupting the CIFAR-10 dataset~\cite{ref_cifar}. Consequently, this dataset is annotated with category of object and type of corruption used. Each object class is highly correlated with a certain type of corruption. We select the object type and corruption type as the target and bias attributes, respectively. We conduct experiments by varying the ratio of \textit{bias-conflicting} samples by selecting from \{0.5\%, 1.0\%, 2.0\%, 5.0\%\}. 

\paragraph{Biased FFHQ} was proposed by~\cite{kim2021biaswap} and curated from FFHQ datset~\cite{karras2019style}. It contains human face images that annotated with the gender and age. In the bFFHQ, 99.5\% of the women are young (age: 10-29), and 99.5\% of the men are old (age: 40-59). Therefore, the ratio of the \textit{bias-conflicting} samples of the bFFHQ is 0.5\%. We select the age as the target attribute, and the gender as the bias attribute.

\paragraph{Celeb-A} is a large-scale face attributes dataset~\cite{liu2015deep}. It contains a total of 202,599 images annotated with 40 binary attributes and 5 landmark location. Following the official train-test split, we train with 162,770 training images and evaluate accuracy on 19,962 test images. Following~\cite{nam2020learning}~\cite{kim2022learning}, we select \textit{BlondHair}, \textit{HeavyMakeup}, \textit{Attractiveness}, \textit{Bignose}, \textit{Bag-under-eyes}, \textit{Male}, and \textit{Young} as the attribute candidates, and choose the highly correlated target and bias attributes among them. For example, in Celeb-A dataset, the most of the male images doesn't have a blond hair or a heavy makeup.

\paragraph{WaterBirds}~\cite{sagawa2019distributionally} is a dataset in which the target attribute is bird species (waterbird vs. landbird) and the bias is the background (water or land). WaterBirds dataset is constructed by combining bird with backgrounds in a biased way: most waterbirds are pictured on water (\textit{bias-aligned}) and most landbirds on land, while only a 5\% of images are placed in the opposite background (\textit{bias-conflicting}).

\section{Implementation Details}

\paragraph{Training Configuration.} We follow the training settings of the previous works, LfF~\cite{nam2020learning}, DisEnt~\cite{lee2021learning} and CSAD~\cite{zhu2021learning}. We employ Pytorch \texttt{torchvision} implementations of the ResNet-18 and ResNet-50 as the encoder network. We train the network with Adam optimizer with the default parameters ($\beta_1=0.9$ and $\beta_2=0.999$) and weight decay of 0. We set the decaying step to 10K. For Celeb-A and WaterBirds, following previous works~\cite{nam2020learning, tartaglione2021end, zhu2021learning}, we employ ImageNet pretrained ResNet-18 and ResNet-50 which are provided by Pytorch \texttt{torchvision}. Experiments were run on NVIDIA Titan Xp GPUs.

\section{Evaluation Metrics}
Let $(\hat{Y},Y,B) \in \mathcal{Y} \times \mathcal{Y} \times \mathcal{B}$ denote the prediction, target attribute, and bias attribute, respectively. The \textit{worst-group} test accuracy can be expressed using the following equation: 
\begin{align}
    acc_{wg} =  \min_{y,b\in\mathcal{Y}\times \mathcal{B}} p(\hat{Y} = y|Y=y,B=b).
\end{align} Then, we can express the \textit{unbiased} and \textit{bias-conflicting} test accuracy using the following equation:
\begin{align}
    acc = \frac{1}{\lvert \Omega \rvert }\sum_{(y,b)\in \Omega} p(\hat{Y} = y|Y=y,B=b),
\end{align}
where $\Omega$ is the set of the target-bias pairs. To calculate the \textit{unbiased} test accuracy, we average the test accuracy over the all possible target-bias pairs (i.e., $\Omega=\mathcal{Y} \times \mathcal{B}$). To calculate the \textit{bias-conflicting} test accuracy, we average the test accuracy over the \textit{bias-conflicting} target-bias pairs (e.g., old-woman, young-man in bFFHQ, and \textit{BlondHair} Male, not \textit{BlondHair} Female in Celeb-A). 

\section{Results on WaterBirds}
\begin{table}[t]
\centering
\resizebox{0.5\columnwidth}{!}{
\begin{tabular}{lccccc}
\toprule
Method   & Backbone & Bias Label     & \textit{Worst-G}   &  \textit{Mean}   \\ \midrule
Vanilla  & Res50 & \xmark & 74.9\stdv2.4 & \textbf{98.1\stdv0.1}     \\
LfF~\cite{nam2020learning} & Res50 & \xmark & 78 & 91.2 \\
JTT~\cite{liu2021just} & Res50 & \xmark & 86.7 & 93.3 \\
SSA~\cite{nam2022spread} & Res50 & $\triangle$ & 89.0\stdv{0.6} & 92.2\stdv{0.9} \\
CNC~\cite{zhang2022correct} & Res50 & $\triangle$ & 88.5\stdv{0.3} & 90.9\stdv{0.1} \\
\DFRVAL~\cite{kirichenko2022last} & Res50  & $\triangle$ & \underline{92.9{\stdv0.2}} & 94.2{\stdv0.4} \\
GroupDro~\cite{sagawa2019distributionally} & Res50  & \cmark &  91.4 & 93.5  \\
LISA~\cite{yao2022improving} & Res50  & \cmark &  89.2 & 91.8  \\
Ours    &  Res50 & \cmark    &  \textbf{93.46}{\stdv0.11}   &  \underline{96.82{\stdv0.52}} \\ \midrule 
Vanilla  & Res50(w/o IN) & \xmark & 6.9\stdv3.0 & \textbf{88.0\stdv1.1}     \\
\DFRVAL~\cite{kirichenko2022last} &  Res50(w/o IN) & $\triangle$ & \underline{56.70\stdv1.3} &  61.03\stdv1.62  \\ 
Ours &  Res50(w/o IN)& \cmark & \textbf{63.56\stdv0.76} &   \underline{69.72\stdv1.14}
 \\ \bottomrule 
\end{tabular}
}
\caption{Comparison of the \textit{Mean} and \textit{Worst-Group} test accuracy (\%) on the WaterBirds. Best performing results are marked in bold, while the second-best results are denoted with underlines. Res50(w/o IN) referes to ResNet-50 without ImageNet pretraining.
}
\label{tab:waterbirds}
\end{table}

In Table~\ref{tab:waterbirds}, we report the \textit{Mean} and \textit{Worst-Group} test accuracy on WaterBirds~\cite{sagawa2019distributionally}. By following~\cite{sagawa2019distributionally, kirichenko2022last}, we compute \textit{Mean} accuracy by weighting the group accuracies according to their prevalence in the training data. For \textit{Worst-Group} accuracy, our method consistently outperforms all competing algorithms, demonstrating its effectiveness in mitigating bias. For \textit{Mean} accuracy, our method achieves the best performance after \textit{Vanilla}, which performs well on the \textit{bias-aligned} samples but fails on \textit{bias-conflicting} samples. These results confirm that our approach achieves both superior fairness and overall performance. Notably, the performance gap between our method and DFR, the second-best performing approach, increases when the backbone network is not pretrained on ImageNet. This suggests that despite the need for bias labels, our method remains more robust and broadly applicable across different settings.

\end{document}